\documentclass[journal]{IEEEtran}
\ifCLASSINFOpdf
  \usepackage[pdftex]{graphicx}
\else
\fi
\usepackage{amsmath}
\usepackage{amssymb}

\DeclareMathOperator*{\argmin}{argmin}

\begin{document}
%
\title{Recent Progress in Appearance-based Action Recognition}
%
%
%

\author{Jack~Humphreys, Zhe~Chen, and~Dacheng~Tao ~\IEEEmembership{Fellow, ~IEEE}
\IEEEcompsocitemizethanks{
\IEEEcompsocthanksitem J. Humphreys and Z. Chen have contributed equally. 
\IEEEcompsocthanksitem J. Humphreys, Z. Chen, and D. Tao are with the UBTECH Sydney AI Centre, the University of Sydney, Darlington, NSW, 2008, Australia. E-mail: jhum4583@uni.sydney.edu.au; zhe.chen1@sydney.edu.au; dacheng.tao@sydney.edu.au
}
}

%
%

\markboth{Journal of \LaTeX\ Class Files,~Vol.~xx, No.~xx, November~2020}%
{Jack \MakeLowercase{\textit{et al.}}: Recent Progress in Appearance-based Action Recognition}
%



\maketitle

\begin{abstract}
Action recognition, which is formulated as a task to identify various human actions in a video, has attracted increasing interest from computer vision researchers due to its importance in various applications. Recently, appearance-based methods have achieved promising progress towards accurate action recognition. In general, these methods mainly fulfill the task by applying various schemes to model spatial and temporal visual information effectively. To better understand the current progress of appearance-based action recognition, we provide a comprehensive review of recent achievements in this area. In particular, we summarise and discuss several dozens of related research papers, which can be roughly divided into four categories according to different appearance modelling strategies. The obtained categories include 2D convolutional methods, 3D convolutional methods, motion representation-based methods, and context representation-based methods. We analyse and discuss representative methods from each category, comprehensively. Empirical results are also summarised to better illustrate cutting-edge algorithms. We conclude by identifying important areas for future research gleaned from our categorisation.
\end{abstract}


%
\IEEEpeerreviewmaketitle

\section{Introduction}

\begin{figure}
    \centering
    \includegraphics[width=0.5\textwidth]{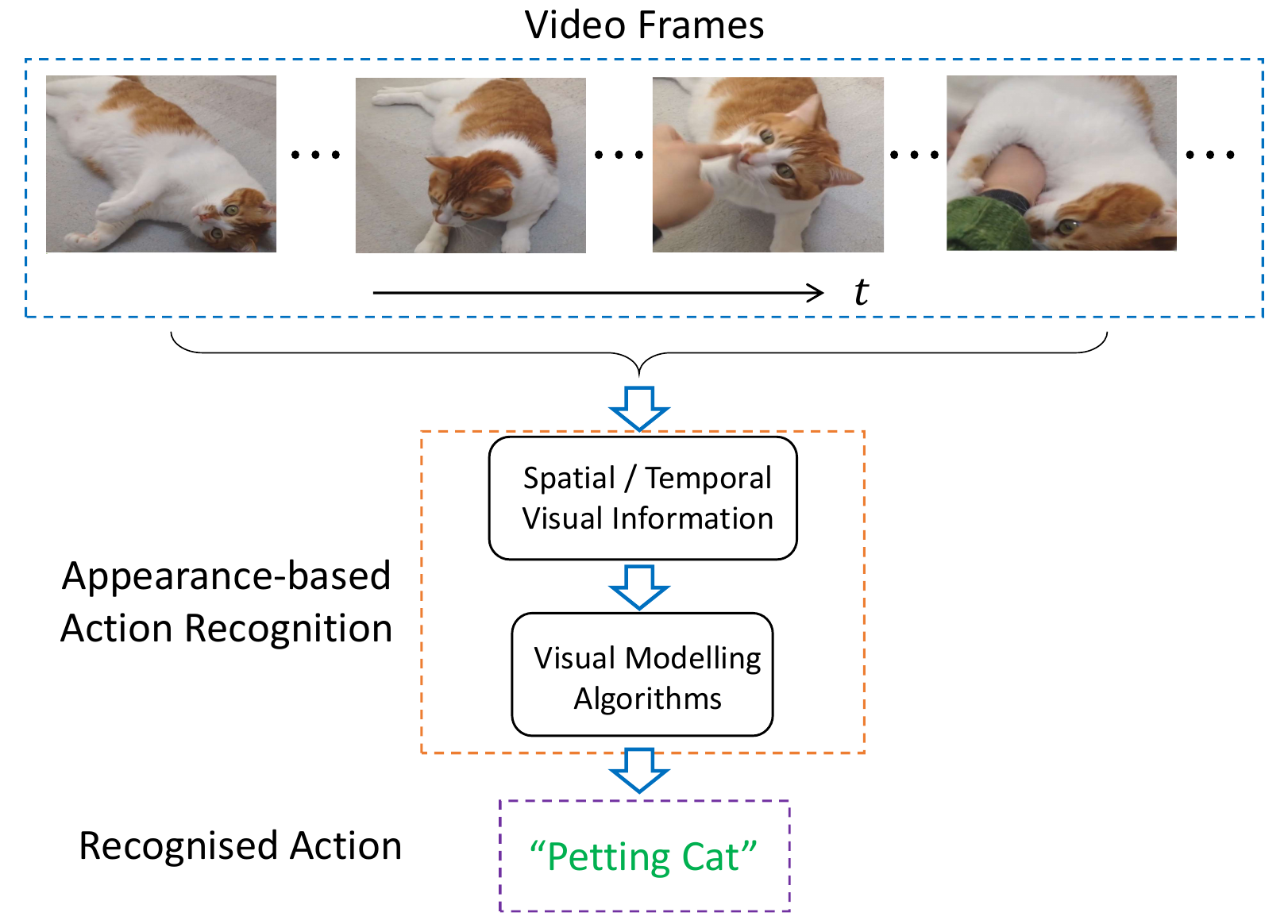}
    \caption{A brief example from \textbf{Kinetics400} dataset \cite{kay2017kinetics}, illustrating the general processing logic of appearance-baesd action recognition methods. Given a video with frames stacked across time axis $t$, appearance-based action recognition methods model spatial and temporal visual information and then predict the action appeared in the video based on the modelled appearance information.
    }
    \label{fig:examples}
\end{figure}

Action recognition (AR) is usually formulated as a task that aims to recognise the primary human action being performed within a given video. Despite the constraint that such videos mainly feature human actions, AR is integral to achieving generic video understanding and benefiting many downstream video understanding applications, such as video retrieval, self-driving systems, surveillance, robotics, \textit{etc.}. In AR, involved actions could be `knitting', `playing flute', `haircut', and so on. Figure \ref{fig:examples} shows a brief example in which a human is petting a cat in the video. 
In general, action classification, which assigns a label to a video to indicate the occurred actions in the video sequence, is the abstract and foundation of various action recognition down-stream tasks like action detection and action localisation. In this paper, we mainly study the action classification approaches for action recognition.

\begin{figure*}
    \centering
\includegraphics[width=\textwidth]{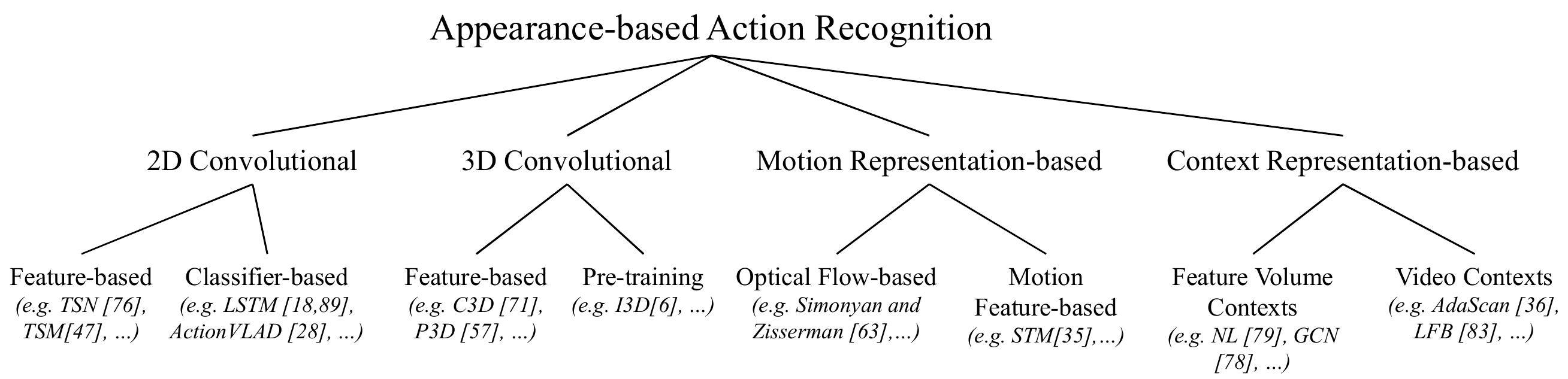}
\caption{Tree diagram for different categories and subcategories of appearance-based action recognition methods.}
\label{fig:categorisation}
\end{figure*}
Over recent years, researchers who study the action recognition problem have widely applied deep neural networks (DNNs) for action recognition. DNNs, especially deep convolutional neural networks (DCNNs), have facilitated researchers to achieve great success in various image recognition tasks, as well as action recognition, due to their impressive representational power and large annotated datasets available for training.
Existing DNN-based action recognition algorithms either exploit human pose information to estimate actions or purely perceive actions based on visual appearances in the video. Pose-based methods first estimate human pose, \textit{i.e.} keypoints indicating the positions of the head, shoulder, \textit{etc.}., from videos, and then recognise actions based on estimated poses. Correspondingly, appearance-based methods utilise RGB colors of one or several frames of the video as their primary input modality and recognises the action based on this color information. Although both types of methods have made promising progress, appearance-based methods are attracting more interest from researchers because pose-based methods usually perform poorly when only RGB information is available\cite{yan2018spatial}. For example, Choutas \textit{et al.} \cite{choutas2018potion} showed that their successful pose-based method degraded the performance of a simple appearance-based architecture in the absence of depth data and where the subject is partially occluded. To this end, in this survey, we put our focus on the appearance-based action recognition algorithms. By excluding pose-based action recognition methods, subsequent references to ``action recognition'' within the remainder of this article are synonymous with ``appearance-based action recognition''.

Since appearance-based action recognition methods classify videos into actions mainly based on visual information, how to better describe appearances is one of the most important questions for solving the problem of action recognition. To answer this question, researchers have introduced various strategies that differ from each other in the way how representations, or features, of the video data are learned and extracted. According to our study, existing appearance-based action recognition methods can be roughly grouped into four main categories based on their mechanisms to extract features: 1) \emph{2D convolutional methods} that mainly extract features from still images with 2D convolutions; 2) \emph{3D convolutional methods} that extract features from a series of subsequent frames with 3D convolutions; 3) \emph{Motion-based methods} that particularly extract motions from videos to help tackle the action recognition problem; and 4) \emph{Context-based methods} that tend to extract contextual features from surrounding areas or frames to benefit the recognition of actions.


2D convolutional methods employ the common 2D DCNNs to spatially represent actions within static frames that, when aggregated over time, achieve a classification for the whole video. These action representations either utilise interactions across the features from subsequent frames or are used directly by a classifier. In cases where actions can be recognised by purely intra-frame features, simple temporal aggregation techniques, such as mean or max pooling across all sampled frames, are sufficient for effective action recognition. 
2D convolutional methods can take advantage of the representational power of cutting-edge 2D DCNNs, but they require the additional facilitation of temporal interactions in the feature learning process \cite{lin2019tsm} or the more complicated classifiers that subsume temporal reasoning \cite{donahue2015long} for successful recognition.

3D convolutional methods exploit the use of 3D convolution which extends the 2D convolution by further convolving over an extra time dimension. The time dimension is obtained by stacking adjacent frames across time. Different from 2D convolutional methods which convolve individual frames and aggregate the outputs, 3D convolutional methods directly encode spatial and temporal information within features, thereby achieving spatio-temporal representations of actions.
Using inter-frame information from the outset is beneficial to recognize the semantics of actions contained within the video sequence rather than the frames themselves, and 3D convolutions are able to model this inter-frame information within their 3D kernel. However, 3D kernels can be difficult to learn because inter-frame transformations can be very complex, non-affine, and have varied temporal extents. The complexity of the 3D kernel, both in terms of computational cost and difficulty learning, has motivated many methods including decomposing the 3D kernel into separate spatial and temporal convolutions \cite{feichtenhofer2016spatiotemporal,qiu2017learning,xie2018rethinking,tran2018closer} and initialising the 3D kernel with inflated spatial features \cite{carreira2017quo}. 

Besides the 2D or 3D convolutional methods, researchers also utilized motion information as a powerful inductive bias for representing appearances. In particular, different from 3D convolutional methods, motion information can be usually obtained based on low-level, affine transformation, such as optical flow, to describe the temporal changes of visual contents in a video. Such motion information provides beneficial cues for describing an action but can be fraught with challenges, especially considering the computation costs of traditional ways to extract motion cues \cite{zach2007duality}. Alternative to traditional ways, many current motion-based representations introduce an additional CNN to learn motion features from various input modalities \cite{simonyan2014two,wu2018compressed}.


\begin{figure*}
    \centering
    \includegraphics[width=\textwidth]{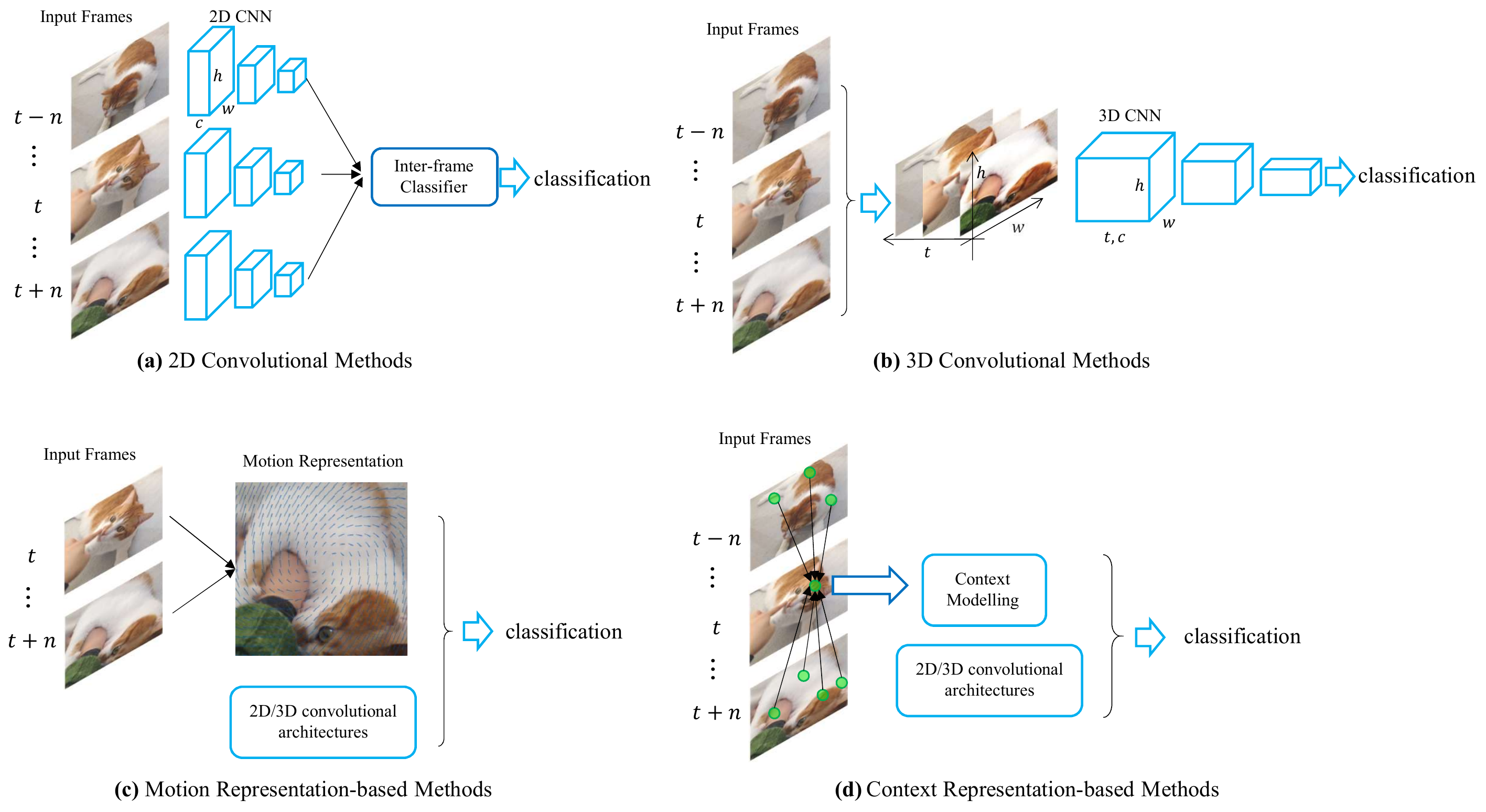}
    \caption{Brief illustration for our categorisation of appearance-based action recognition methods. 
    }
    \label{fig:infographic}
\end{figure*}
Contexts have been shown to be advantageous for various computation tasks \cite{chen2020recursive,chen2017spatial,zhao2017pyramid}, including action recognition. Context features can be used to augment individual features or emphasise them by including the relationships to other features. Context representations for action recognition can be either the global video context in order to make sure key frames featuring the action are sampled \cite{kar2017adascan,wu2019liteeval,wu2019adaframe} or the global feature volume context to facilitate higher level reasoning across a wide spatio-temporal extent for representing actions \cite{wang2018non}. Despite the benefits, the extraction of the context features can be more time-consuming in pursuing higher accuracy. 


In summary, our categorisation can be summarized in figure \ref{fig:categorisation} with a detailed explanation in the following section. In the rest of this paper, we will discuss each category of reviewed methods providing comparisons within the context of our categorisation. For example, Section \ref{sec:2d} and Section \ref{sec:3d} describe the 2D convolutional methods and 3D convolutional methods, respectively. We then review motion-based methods in Section \ref{sec:mot} and context-based methods in Section \ref{sec:ctx} subsequently. In Section \ref{sec:exp}, we will briefly analyse the quantitative performance of each of these methods and attempt to draw conclusions as to promising directions for future appearance-based action recognition research.

\section{Overview}

\subsection{Problem Definition} 
We expound our definition of action recognition and our categorisation methodology with the following definitions. In general, action recognition methods specify the function $\mathcal{F}$ which takes as input an ordered set $X$ of sampled frames from an input video to estimate the class $y$ of the action presented in the video: 
\begin{equation}
    y = \mathcal{F}(X, \mathcal{W}),
    \label{eq:def}
\end{equation}
where the function $\mathcal{F}$ is parameterised by $\mathcal{W}$.
Given the ground-truth action class $\hat{y}$ for each video, action recognition can then be formulated as a learning problem described by
\begin{equation}
    \argmin_\mathcal{W} \mathcal{L}(y, \hat{y}),
    \label{eq:loss}
\end{equation}
where $\mathcal{L}$ is some loss function representing the semantic deviation of the recognised action $y$ and the true action class $\hat{y}$. 
Training regimes that implement this problem enable action recognition methods to realise a parameterisation $\mathcal{W}$ with satisfactory estimations of the action class of any video for testing or inference. 
In practice, $y$ and $\hat{y}$ are represented by non-negative integers, each of which describe an action like laughing or running. 
According to different strategies to model temporal or spatial features from visual appearances in $X$, we categorize current deep learning methods for appearance-based action recognition  into four major types. 

\subsection{Method Formulation}
The first type of methods, \textit{i.e.} 2D convolutional methods, tend to focus on the modeling of the spatial information of each frame from $X$. 
This type of methods generally process the the sampled individual frames from $X$ separately. Consider that $X$ has sampled $n$ subsequent frames from the video:
$X = \{x_1, x_2, \dots , x_n\}$,
where $x_i$ represents a single frame in $X$ and $i$ spans the total number of frames within the sampled video sequence. 
Then, by using the form of equation \eqref{eq:def}, 2D convolutional methods can be described by the following function:
\begin{equation}\label{eq:spatial}
    y = \mathcal{F}((x_1, x_2, \dots , x_n),\mathcal{W}) = G(\bigoplus_{i=0}^{n} \mathcal{F}_{2D}(x_i, \mathcal{W}_{2D}), \mathcal{W}_G),
\end{equation}
where $\mathcal{F}_{2D}$ represents the function that models spatial visual appearances of each frame, $\bigoplus$ is some aggregation operation like concatenation, and $G$ is a function that maps the aggregated information to the desired output. 
The function $\mathcal{F}_{2D}$ is commonly implemented by a 2D CNN.
The functions $\mathcal{F}_{2D}$ and $G$ have corresponding parameters $\mathcal{W}_{2D}$ and $\mathcal{W}_G$ such that $\mathcal{W} = \mathcal{W}_G \cup \mathcal{W}_{2D}$.  Note that, in some cases, $G$ is not parameterised such that $\mathcal{W}_G = \emptyset$. 

Although some real world action classes can be directly recognised by the static content of an individual frame, there many cases in which temporal information is important for successful recognition. 3D convolutional methods propose to apply 3D convolutions on the $X$ to obtain temporal information and to facilitate inter-frame reasoning. Comparing to the 2D convolution which operates on images with two spatial dimensions, the 3D convolution operates on the tensor formed by the stack of several subsequent frames form. Such stacked frames result in an additional time dimension. As a result, the spatio-temporal representation of actions can be modeled and encoded within the output features implicitly. A mathematical description of 3D convolutional methods can be described as:
\begin{equation}\label{eq:3d}
    y = \mathcal{F}((x_1, x_2, \dots , x_n),\mathcal{W}) = \mathcal{F}_{3D}(\bigoplus_{i=0}^{n}x_i, \mathcal{W}_{3D}),
\end{equation}
where $\mathcal{F}_{3D}$ represents a 3D CNN parameterized by $\mathcal{W}_{3D}$. 

Motion-based representations, such as the displacement of pixels occurring between adjacent frames, can be a useful and explicit inductive bias for action recognition. 
This inclusion is typically fulfilled by the two-stream architecture framework which is described by the function:
\begin{equation}\label{eq:motion}
    y = G(\mathcal{F}_{2D/3D} (X, \mathcal{W}_{2D/3D}), \mathcal{F}_{M} (R(X), \mathcal{W}_{M})), 
\end{equation}
where the $\mathcal{F}_{2D/3D}$ parameterised by $\mathcal{W}_{2D/3D}$ denotes either 2D CNN or 3D CNN for modeling visual appearance, and $\mathcal{F}_{M}$ parameterised by $\mathcal{W}_{M}$ represents the function that extracts motion information. 
The function $G$ aggregates classifications from the two streams with operations like simple averaging. $R$ denotes the transformation of input sample $X$ into a representation of the motion in $X$ which is then further processed into a motion representation for action recognition by the motion stream $\mathcal{F}_{M}$. 

Lastly, context representation-based methods tend to incorporate additional useful information to complement the representation extracted with normal operations, aiming to improve the recognition performance of complex actions. In particular, complex actions are the actions 
composed of many high-level subactions that can make it difficult to perform accurate recognition without representations of subactions and having these representations be accessible by the same model. An example of this is making tea where many subactions, such as `turning on kettle', `adding teabag' and `pouring', occur over a wide temporal extent. Sparse sampling of frames then makes typical modeling of $X$ difficult to obtain sufficient knowledge about whether the video contents present the action of making tea because some important cues may be ignored.
To this end, the global video representation can be the contexts to help ensure successful recognition. To sum up, the context-based method can be formulated with following function:
\begin{equation}\label{eq:context}
    y = \mathcal{F}_{2D/3D} (\mathcal{F}_{C} (X, X', \mathcal{W}_{C}), \mathcal{W}_{2D/3D}),
\end{equation}
where $\mathcal{F}_{C}$ is the function, optionally parmeterised by $\mathcal{W}_C$, which either learns to augment the input sample $X$ with the video context computed using the input video $X'$ or simply returns the sample for feature volume context methods where the function $\mathcal{F}_{2D/3D}$ includes global connectivity.

For ease of understanding, the summary and detailed categorization of the reviewed methods are visualised in figure \ref{fig:infographic}.


\subsection{General Discussion}

Considering the rapid development of 2D CNNs in computer vision, it is straight-forward for 2D convolutional methods to apply cutting-edge 2D CNNs for action recognition. For example, 2D convolutional methods can enjoy the explicit transferability from image recognition by exploiting thoroughly researched architectures with pre-training on large image recognition datasets like ImageNet \cite{russakovsky2015imagenet}. 2D convolutional methods can also take advantage of very efficient designs of 2D CNNs like \cite{tan2019efficientnet,tan2019efficientdet} to reduce the computational costs for recognizing actions. However, the model capacities of these 2D convolutional methods could be limited because 2D CNNs are commonly optimized according to the recognition based on spatial visual cues from still images. They generally lack the ability to model temporal information comprehensively. 

On the contrary, the power of 3D convolutional methods is undeniable from a purely quantitative performance perspective. This performance may be explained to some extend by a number of advantages 3D convolutional methods enjoy. Firstly, they have much higher capacity due to the additional convolution dimension parameters. This advantage can be significantly useful with the development of much more difficult datasets such as those with very little representation bias or security applications where subjects are adversarially masking their own action. Secondly, 3D convolutions are a relatively generic operation that can be customised, such as by changing the receptive field, to accommodate future advances in deep learning as a whole. The inductive bias that motivates convolution over a given dimension also applies to the temporal dimension. Despite these advantages, the major problems of 3D convolutional methods include the considerable computational costs and the great difficulty for optimizing 3D CNNs. In particular, by introducing an additional dimension, the clip-based nature of 3D methods typically results in sampling vastly more frames in total. Some 3D methods samples 10 times more frames than comparable 2D methods which directly increases computation costs by that same factor ignoring the further costs associated with the larger kernel. Besides, the lacking large-scale video datasets for the pre-training of 3D CNNs and the extra model dynamics accompanied by larger model capacities would pose significant difficulty for training 3D CNNs properly.

In addition to 2D/3D convolutional methods which focus on the backbone network for action recognition, motion-based and context-based methods tend to better achieve the recognition task by introducing extra auxiliary information. Motion-based methods provide particular advantages to architectures by including motion information such as optical flow inherited in videos. This inclusion can take place as an auxiliary motion stream which is entirely supplementary and as such is entirely compatible with other methods. Although spatio-temporal representation methods such as 3D convolutional methods already have the capacity of subsuming motion representations, motion-based studies often experience additional performance gains by incorporating a motion stream. In practice, however, the introduction of motion-based representation is usually at the cost of doubling the computational loads because it usually requires an additional backbone architecture and a pre-processing step to transform the temporal visual information to a preliminary representation of its motion. The introduced computational loads may also make it infeasible for some applications which need real-time processing speed.



Context-based methods usually introduce the global representation of the entire video to extract extra visual cues for improving action recognition. These methods can reason across very wide spatio-temporal receptive field with the potential to encompass the entire volume in a given layer. This can be achieved even in the early layers of the network, enabling features with long-term dependencies to be extracted earlier. The extraction of contexts usually comes at only modest computational costs and can complement the representation of many actions of interest which are difficult to describe with sampled frames only. 
In addition to simply augmenting features, contexts can also be configured to massively reduce the number of frames required to recognise actions by learning to sample salient frames. This also has a further advantageous side effect of using context to emphasise these salient frames improving performance. Both forms of contextual information are compatible other action recognition methods including those from other methods and subsequent backbone CNNs either as a modular residual layer or as simply a function that alters the input either adding context or transforming it based on the context.

\section{2D Convolutional Methods}
\label{sec:2d}
Based on the typical single-frame architecture \cite{karpathy2014large}, 2D convolutional methods commonly study how to better exploit temporal information with 2D CNNs for the action recognition task. Based on different strategies of encoding temporal information, current 2D convolutional methods can be further described by two sub-categories. The first group is to enable temporal interaction between frame-level features in the input or at various depths throughout the 2D CNNs. We name this group as feature-based 2D convolutional methods. For the second group of methods, the classification heads, or classifier, of the 2D CNNs are comprehensively studied so that inter-frame reasoning and temporal reasoning can be subsumed at the highest level of abstraction. A brief illustration about some typical methods of these subcategories are shown in table \ref{tab:subcat2d} for clarity.


\begin{table}
    \begin{center}
    \resizebox{\columnwidth}{!}{%
    \begin{tabular}{rl}
        \hline
        \textbf{Feature-based} & \textbf{Main Contributions}\\
        \hline
        TSN \cite{wang2016temporal} & Uniform sampling, multimodal Video Consensus \\
        MDI \cite{bilen2016dynamic} & 2D video summarisation of input \\
        STCB \cite{wang2017spatiotemporal} & Bilinear fusion of streams \\
        TSM \cite{lin2019tsm,chen2018biglittle,sudhakaran2020gate} & Temporally shifting channels \\
        \hline
        \textbf{Classifier-based} & \textbf{Main contributions}\\
        \hline
        LSTM \cite{donahue2015long,yue2015beyond,sun2017lattice,sudhakaran2019lsta} & LSTM classifier\\
        ActionVLAD \cite{girdhar2017actionvlad} & Closest learnt descriptor with residual vector\\
        SVMP \cite{wang2018video} & SVM hyperplane descriptor\\
        \hline
    \end{tabular}
    }
    \end{center}
    \caption{Subcategories and summary of some typical 2D convolutional methods.}
    \label{tab:subcat2d}
\end{table}

\subsection{Feature-based Methods}

Feature-based methods were pioneered with various forms of temporal fusion including the simplest form of single-frame processing \cite{karpathy2014large}. Later studies then focus on the more specialised feature extraction techniques and more effective training regimes.


The very typical work authored by Karpathy et al. \cite{karpathy2014large} comprehensively investigate the performance of CNN architectures for large-scale action recognition using the first dataset of its kind which they introduce therein. The authors compare a variety of architectures for  encoding temporal features with 2D CNNs, including the single frame-based model, the model with late fusion, the model with early fusion, and the model with slow fusion. Each of these architecture recognises actions based on randomly sampled clips of contiguous frames with clips being averaged for test time video-level classification. In this study, it has shown that the performance is not very sensitive to the architectural details of encoding visual changes along time, even for the sport videos with more dynamic appearances.


Following \cite{karpathy2014large}, researchers have tried to develop more effective feature extraction architectures to model connectivity along time for action recognition. For example, there exists a group of methods which perform feature extraction that is capable of modelling temporal interactions either in the input or in latent space as with the early fusion and slow fusion, respectively. Bilen et al. \cite{bilen2016dynamic} pool video features into a single `dynamic' image enabling powerful 2D CNNs to be directly applied to videos. This is achieved through approximate rank pooling which the authors extended to be applicable to feature maps as well as videos represented by other modalities such as optical flow. Wang et al. \cite{wang2017spatiotemporal} extend feature extraction methods to two-stream architectures by using cross-stream correspondences to model temporal interactions. These correspondences are extracted hierarchically using bilinear fusion. 
Lin et al. \cite{lin2019tsm} propose a fully 2D decomposition of 3D convolutions which involves temporally shifting a portion of channels and folding the multiplication and accumulation operations for the time dimension into the one that is performed along the channel dimension. Specifically, a fraction of the channels in each layer are shifted by one voxel in the time dimension, half forwards, half backwards. 
The resulting features outsource temporal interactions to the channel dimension as with slow and early fusion methods. 
Fan et al. \cite{fan2019more} extend the receptive field of these temporal shifts using the Big-Little Net image recognition model \cite{chen2018biglittle}. This model features one stream with a high channel capacity which operates at a low spatial resolution and another stream operating at a high resolution but without as many channels. The features from the two streams are aggregated throughout the network allowing interaction between the streams. 
To extend the limited receptive field, the authors propose a module which is interleaved with convolutional layers and entails weighting feature channels with depth-separable convolutions and aggregating neighbouring channels within a hyperparameterised temporal window. 
Sudhakaran et al. \cite{sudhakaran2020gate} propose a different extension to temporal shifting where the shifting mechanism is discriminitvely learnt. This is achieved using adaptive gating which learns to shift channels forward or backward in time where the modes of the activation function computing the average or difference of shifted channels. 

Derived from \cite{karpathy2014large}, another group of methods implicitly refine the learnt features to emphasise temporal reasoning which is otherwise lacking in 2D CNN architectures. Wang et al. \cite{wang2016temporal} propose to capture long-term temporal dependencies by sparsely sampling snippets uniformly and by providing a video level supervisory signal rather than a snippet level one. Snippets, in this case, are composed of a stack of 5 frames for the motion stream within their two-stream architecture, along with a single frame for the appearance stream. 
Roberto de Souza et al. \cite{roberto2017procedural} observe that there is more space to utilise procedurally generated data which can emphasise the temporal evolution of the appearance of frames. The authors validate that synthetic video data can impose regularising effects when trained jointly with small real world datasets. Rather than implicitly learn temporal interaction with the supervisory signal or selection of data, Fernando et al. \cite{fernando2017self} propose to set an explicit, temporally focused self-supervised learning task, odd-one-out. This is where a 2D CNN learns to correctly ascertain which, of a number of videos, has had the ordering of its frames perturbed. This method encourages the learning of meaningful semantic features as it uses both positive and negative examples each iteration. Lin et al. \cite{lin2018action} propose a framework that learns coarse-to-fine grain representations using early prediction losses on coarse grain feature maps as well as coarse grain meta-class losses with the ground-truth being approximated by a lightweight CNN. 

\subsection{Classifier-based Methods}

Rather than feature-based methods, classifier-based methods generally investigate effective mechanisms to model action across time with the most abstract visual features extracted with 2D CNNs. There are two main groups of classifier-based methods. The first group all use classifiers capable of temporal reasoning such as long short-term memory (LSTM) from the RNN family of neural networks. The second group assume that actions can be uniquely identified by appearance and/or motion features and, thus, do not require specific temporal reasoning for classification. These methods opt to directly exploit this inductive bias in the design of the classifier.

The first method to utilize LSTM is that of Donahue et al. \cite{donahue2015long} who enable the modelling of long-term spatio-temporal dependencies of actions. They first compute two-stream CNN features for each frame in a given clip. Then, they process the 2D CNN features sequentially using an LSTM with the final clip level prediction being the average of its outputs at all time steps. Subsequent methods increase the capacity of this LSTM classifier. One such method is proposed by Yue-Hei Ng et al. \cite{yue2015beyond} which performs spatial max pooling over CNN feature maps for the whole video and aggregates these features using a stack of five LSTMs. The final classification is the linearly weighted average of the LSTM stack outputs such that the output for the first frame has 0 weight and the last has 1. 
The authors also propose a novel training technique whereby the bulk of the training can be done on very few or even a single frame with fine-tuning being done at progressively higher numbers of frames. Sun et al. \cite{sun2017lattice} claim that these methods do not allow modelling of long range dependencies in the spatial dimensions due to local patching and max pooling, respectively. To alleviate this, the authors propose the Lattice LSTM ($\textrm{L}^2$STM) which learns separate state transition functions for each pixel location and facilitates inter-pixel interactions with a recurrent attention layer. Sudhkaran et al. \cite{sudhakaran2019lsta} are motivated by the fact that videos are biased towards consistent movement across time to propose the long short-term attention classifier. This modified version of an LSTM applies spatio-temporal attention to the input features using learnt latent category filters which are processed into the attention map by an additional RNN. The category filters are pooled over in order to track the most important latent categories, such as types of object, across space and time.

A further group of methods utilise inductive biases in the form of video level descriptors which isolate the key discriminitive semantics in high level feature maps for classification. Girhar et al. \cite{girdhar2017actionvlad} consider the semantics of an actions as a set of high level appearance and motion feature descriptors/subactions. This concept is related in the form of an example of the `basketball game' action class which can be recognised by running and jumping subactions as well as appearance descriptors for basketballs and hoops. The authors introduce this inductive bias to two-stream action recognition methods via the proposed ActionVLAD which is the conceptual extension of NetVLAD \cite{arandjelovic2016netvlad} to the temporal dimension. 
In another study, Wang et al. \cite{wang2018video} observe that methods of aggregating frame level features over time overly bias recognition towards appearance features that are commonly between many classes. They assume that there exists at least one frame whose appearance features enable correct recognition.
Under these assumptions, successful recognition requires disentangling the key feature of this frame from the redundant remainder of the video. In practice, the authors learn an SVM hyperplane to separate this unknown feature from the rest and use the parameters of this hyperplane as a descriptor for the entire video thereby pooling over all the features. This hyperplane is learnt discriminatively and thus can train end-to-end.

\subsection{Discussion}

Despite that both the surveyed feature-based methods and classifier-based methods show promising results for action recognition, the major difference of both groups is the depth at which this reasoning is performed.  
Feature-based methods are able to perform temporal reasoning at multiple levels, or depths, of abstraction within the CNN feature hierarchy before the spatial dimension are marginalised out. They are also able to refine their CNN descriptors in conjunction with added temporal reasoning. This has more potential to exploit the full advantage of the backbone 2D CNN for action recognition, but feature-based methods would require extensive retraining to accommodate. Conversely, classifier-based methods focus on the layers of the classifier or, for recurrent classifiers, the number of frames per sample. They are compatible with generic frame level 2D CNN descriptors, thus can be benefited from recent successful progress of 2D CNNs for image recognition. These methods also have the potential for greatly reduced computational costs. A drawback of classifier-based methods is that they use much shallower temporal reasoning with the depth being limited to the employed classifiers.


\section{3D Convolutional Methods}
\label{sec:3d}
Comparing to 2D convolutional methods, 3D convolutional methods study different strategies to improve the effectiveness of 3D CNNs for action recognition. Since 3D CNNs convolve an additional time dimension of input data, corresponding researchers mainly focus on how to better obtain and utilise the temporal information with 3D CNNs. We further categorise 3D convolutional methods into two groups. The first group develops advanced 3D CNN structure for describing temporal information, and the second group studies improved pre-training mechanisms to improve performance for kernels in 3D CNNs. We name the first group and the second group as feature-based methods and initialisation-based methods, respectively. A brief description of both groups of methods can be found in Table \ref{tab:subcat3d}. 


\begin{table}
    \begin{center}
    \resizebox{\columnwidth}{!}{%
    \begin{tabular}{rl}
        \hline
        \textbf{Feature-based} & \textbf{Main Contributions}\\
        \hline
        ST-ResNet \cite{feichtenhofer2016spatiotemporal,qiu2017learning,xie2018rethinking,tran2018closer} & 2D spatial and 1D temporal decomposition\\
        AssembleNet \cite{ryoo2019assemblenet} & Neural architecture search\\
        SlowFast \cite{feichtenhofer2019slowfast} & Dual frequency frame sampling\\
        \hline
        \textbf{Pre-training} & \textbf{Main Contributions}\\
        \hline
        I3D \cite{carreira2017quo} & Temporal replication of pre-trained 2D features\\
        IG-Kinetics \cite{ghadiyaram2019large} & Large scale pre-Training \\
        \cite{wei2018learning,dwibedi2019temporal,xu2019self,korbar2018cooperative,kim2019self,wang2019self} & Spatio-temporal self-supervised tasks \\
        \hline
    \end{tabular}
    }
    \end{center}
    \caption{Subcategories and summary of representative 3D convolutional methods.}
    \label{tab:subcat3d}
\end{table}

\subsection{Feature-based Methods}
The feature-based methods of 3D convolutional action recognition algorithms study different forms of 3D convolutional kernels. The first type of feature-based methods investigate fully 3D convolutional kernels as pioneered in \cite{ji20123d,tran2015learning}.
In Tran et al. \cite{tran2015learning}, the authors initiate their successful application to action recognition by using a homogeneous 3D CNN architecture. A large-scale dataset is also introduced for pre-training in order to reduce the degree to which 3D kernels overfit. Hara et al. \cite{hara2018can} analyse this susceptibility by applying various 3D residual CNN architectures to both large and small-scale datasets, finding that the latter indeed induce overfitting. Within the groups of residual architectures examined, trends similar to those of image recognition were observed with group convolutions being part of the best networks.
Luo and Yuille et al. \cite{luo2019grouped} utilise group convolutions with two groups, one made up of 2D convolutions and the other of 3D convolutions. The latter is allocated with a smaller fraction of the output channels to account for the large computational cost of the 3D kernel. The authors analyse the contribution of each of these groups finding that the 2D group's features are emphasised in earlier blocks and the 3D group's in the latter half of the network. In addition, Tran et al. \cite{tran2019video} find that channel interactions are a key determinant of action recognition performance such that only very deep models are able to balance the reduced channel interactions of group convolutions. The authors utilise depth-separable bottleneck 3D convolutions on very deep architectures enabling them to leverage the substantial regularising effect of channel separation. Chen et al. \cite{chen2018multi} propose a group convolutional architecture that uses both 2D and 3D convolutions. Its channel interactions are enabled by a `multiplexer' composed of inter-group point convolutions. This multiplexer operates on a smaller latent subspace and enables global information to be routed and aggregated across groups. Zhou et al. \cite{zhou2018mict} mix 2D and 3D convolutions such that their representations are concatenated and fused by further 2D convolutions. The parallel 2D branch frees up the capacity of the 3D convolutions to focus on more complex representations and balances the overall capacity towards spatial representations thereby accounting for the greater redundancy of temporal information in action recognition videos. Similarly, Wang et al. \cite{wang2018appearance} decompose the computation of appearance and relationship features with a 2D convolutional branch and 3D convolutional branch, respectively. The relation branch squares the 3D convolutional activities to attain inter-patch relationships which then undergo cross channel pooling. Choi et al. \cite{choi2019can} reduce the scene bias in 3D CNN architectures by formulating two auxiliary losses based on the outputs of a pre-trained human figure detector and a pre-trained scene classifier. As an example, the authors use the former's output to occlude the human figure in each frame and penalise confident classifications which must inherently be scene biased.

Besides full 3D convolution, another type of feature-based 3D convolutional methods attempt to decompose 3D convolutions to reduce computational cost and make learning easier. The decomposition commonly transforms 3D convolutions into the sequential application of a 2D spatial convolutions and a 1D temporal convolution. 
Feichtenhofer et al. \cite{feichtenhofer2016spatiotemporal} augment 2D residual CNNs with temporal convolutions with kernel size $>1$ in some of its layers thereby resulting in the decomposed 3D convolutions which are explicitly designed by subsequent methods. Qiu et al. \cite{qiu2017learning} decompose the full, $3 \times 3 \times 3$ 3D convolution kernel into a spatial $3 \times 3 \times 1$ convolution and a temporal $1 \times 1 \times 3$ convolution. The authors examine three possible implementations: applying the spatial and temporal convolutions in parallel, sequentially or sequentially but with features maps from both stages being aggregated. They find that interleaving all three throughout the network performed best but later studies reverted to homogeneous architectures finding those to be superior. One such study is that of Tran et al. \cite{tran2018closer} which also showed that both test and training error was reduced by decomposing 3D convolutions, suggesting that they are easier to optimise rather than them inducing a regularising effect. They also find that the performance of 3D CNNs, relative to 2D, increases as clip length does which motivated them to further improve the performance of their decomposed 3D CNN architecture by fine-tuning them on longer clips. Xie et al. \cite{xie2018rethinking} proposed another decomposed architecture but also find that, when removing temporal convolutions from different layers in the network, architectures which retained them in the layers at the top of the network were superior to those which used them in the early layers. Li et al. \cite{li2019collaborative} proposed a different decomposition where they split the 3D kernel into intersecting planes by processing each combination of dimensions, or `view' of the spatio-temporal feature volume, with a weight shared 2D convolution, and concatenating the resulting feature maps.

In addition to directly decomposing 3D convolutions, many methods propose various methods to refine the features they learn.
Feichtenhofer et al. \cite{feichtenhofer2017spatiotemporal} encourage feature correspondence between the two-streams of 3D CNNs using asymmetric multiplicative gating of the appearance stream with features from the motion stream. Zhao et al. \cite{zhao2018trajectory} claim that decomposed 3D CNNs ought take into account the misalignment of spatial features caused by motion. They propose to realign kernels across time using deformable convolutions. Rather than learning adaptive deformable kernel offsets, the authors use an optical flow field of the input feature maps approximated by a CNN \cite{zhu2018hidden}. Ryoo et al. \cite{ryoo2019assemblenet} investigate the structure of decomposed 3D CNN architectures using neural architecture search with the search space constrained to the connections between modality streams and the temporal resolution of each of these streams in the form of dilated temporal convolutions. The optimisation used for the architecture search is a genetic algorithm which is guided by epigentic principals by weighting stream connection mutations by discriminatively learnt gating parameters. He et al. \cite{he2019stnet} propose to further reduce the computational cost of decomposed 3D convolutional methods by stacking input frames into a super-image of 5 frames thereby subsuming the clip dimension into the channel dimension. To facilitate interaction between the temporal and channel dimensions the authors propose a block consisting of depth-separable convolution and subsequent temporal convolution. Feichtenhofer et al. \cite{feichtenhofer2019slowfast} observe that the inductive bias that motivates 2D convolutions, that both spatial dimensions are isotropic, does not hold for the temporal dimension of 3D convolutions. The authors propose a slow fast dichotomy of spatio-temporal features to align with orientation biases present in human actions. This is implemented by factorising the spatio-temporal modelling of a decomposed 3D CNN into two streams with a low frame rate stream with many channels for slow spatial semantics and a stream for fast motion with a much higher frame rate offset by proportionally fewer channels. Kim et al. \cite{kim2020regularization} propose to regularise decomposed 3D CNNs by perturbing the low frequency signal in action recognition videos which the authors show to be less important to successful recognition than the high frequency signal. This is achieved by 3D average pooling over the feature volume yielding the smoothed low frequency signal which is perturbed by a sampled scalar during training and residually connected to the original feature map with the high frequency signal intact.

\subsection{Pre-training Methods}

Based on 3D CNNs, researchers analyse pre-training mechanisms mainly to provide a pre-training that, after training on action recognition datasets, yields better performance, or to achieve comparable performance to normal supervised pre-training without needing access to any labelled data pre-training data.


In particular, there is a group of pre-training methods that propose alternatives to normal supervised pre-training. Carreira and Zisserman \cite{carreira2017quo} enable the use of image recognition datasets for pre-training 3D convolutional action recognition methods by inflating the weights of pre-trained 2D convolution layers to 3D which involves, simply, replicating weights across the time dimension. This is analogous to being trained on very boring videos where a only single frame is repeated across time. Despite the success of this technique, Ghadiyaram et al. \cite{ghadiyaram2019large} later demonstrated that pre-training 3D networks with video data is vastly superior to pre-training on image data. Their proposed initialisation method was to pre-train action recognition models using a huge dataset of 65 million weakly labelled (tagged) videos which they also split into frame level image samples to facilitate a fair comparison with image-based pre-training. 

In the absence of vast quantities of labelled data, self-supervsied learning presents and alternative means of providing a good pre-training scheme for action recognition models by learning some intrinsic characteristic of unlabelled videos. Wei et al. \cite{wei2018learning} propose the self-supervised learning task of explicitly learn the arrow of time where optical flow is computed and used to classify videos playing forward and videos playing backwards. Korbar et al. \cite{korbar2018cooperative} also introduce another modality in their audio-visual temporal synchronisation learning task. The supervisory signal used by this method is whether a pair of sampled visual and audio clips from a video are sampled over the same time interval or different intervals from the video. Dwibedi et al. \cite{dwibedi2019temporal} propose a self-supervised learning method which involves learning a 3D convolutional embedding space for video frames. The network which performs the mapping is trained using a cycle consistency loss which is zero if, for two given frames from different videos, they are each others nearest neighbours in the embedding space. The authors quantify the level of inconsistency by recasting this loss as a regression task with a Gaussian prior. Xu et al. \cite{xu2019self} investigate the clip order prediction learning task which deviates from existing methods that use the true ordering of random permutations of the input in that the permutations are done to temporally contiguous clips rather than individual frames. Kim et al. \cite{kim2019self} extend this method to include spatial semantics. Here, the supervisory signal is either the correct temporal ordering of clips sampled from a spatial quadrant of a video or the correct spatial configuration of spatial quadrants sampled from a quarter of the temporal extent of the video. By sharing weights across quadrants, the model learns to solve both spatial and temporal puzzles, alternately. Directly modelling brightness through pixel intensities, as is done throughout computer vision, can potentially be fraught with challenging sources or modulators of light other than that of the semantic class the image belongs to. Alayrac et al. \cite{alayrac2019visual} propose to disentangle these confounding layers by training a model with the task of reconstructing two separate videos from their blended, convex combination. Although not investigated by the authors, this task can be recast as a self-supervised learning problem. Wang et al. \cite{wang2019self} propose yet another self-supervised method which is to regress various statistics computed on spatial blocks of the video. The authors split the spatial extent of videos into blocks, pinwheels and concentric squares with each pattern being learnt, concurrently. The statistics regressed for each are the greatest and least of both colour diversity and motion, over the temporal dimension as well as the colour and motion directions themselves for the appearance and motion streams, respectively. 

\subsection{Discussion}
Comparing the two groups of 3D convolutional methods, feature-based methods make efficient use of CNN capacity whereas pre-training methods make the difficult process of learning temporally convolutional representations easier.
Feature-based methods are generally able to better extend model capacity of 3D CNNs for action recognition and can achieve promising results by training from scratch. Pre-training methods, on the other hand, have the potential to take advantage of larger-scale unlabeled data and can be readily applied to future architectural developments for 3D CNNs.

\section{Motion Representation-based Methods}
\label{sec:mot}

\begin{table}
    \begin{center}
    \resizebox{\columnwidth}{!}{%
    \begin{tabular}{rl}
        \hline
        \textbf{Optical Flow-based} & \textbf{Main Contributions}\\
        \hline
        Simonyan and Zisserman \cite{simonyan2014two} & Two-stream architecture \\
        FcF \cite{piergiovanni2019representation} & Second order optical flow \\
        D3D \cite{crasto2019mars,stroud2020d3d} & Distilled optical flow features \\
        \hline
        \textbf{Motion Feature-based} & \textbf{Main Contributions}\\
        \hline
        CoViAR \cite{wu2018compressed} & Model motion features based on compressed videos\\
        OFF \cite{sun2018optical} & Optical flow-guided features\\
        \hline
    \end{tabular}
    }
    \end{center}
    \caption{Summary of representative motion representation methods.}
    \label{tab:subcatmotion}
\end{table}

Rather than encoding temporal information in backbone networks implicitly, there are many papers that propose to extract motion patterns from video explicitly to facilitate the action recognition. The first of such motion representation-based action recognition can be found in 
a two-stream architecture featuring a motion stream \cite{simonyan2014two}. 
Following methods can be roughly described by two types, \textit{i.e.} the ones that extract optical flows and the ones that learn motion features that can be complementary to optical flows. The former ones compute optical flow from subsequent frames to facilitate recognition or use CNN layers to approximate optical flow, enabling motion representations whose input modalities can be optimised for the learning task. The latter ones 
then learn motion features from frames implicitly. We present brief descriptions of some typical methods from this category in Table \ref{tab:subcatmotion}.


\subsection{Optical Flow-based Methods}
Optical flow-based methods can directly compute optical flows to represent motions. For example, Simonyan and Zisserman \cite{simonyan2014two} proposed a two-stream architecture where one CNN is the appearance stream operating on RGB frames and the other is the motion stream whose input is a stack of optical flow frames algorithmically computed using many adjacent RGB frames. 
The algorithm used to compute optical flow here and in all subsequent methods is described in TV-L1 \cite{zach2007duality}. The authors found that optical flow is superior to computing trajectories or
bi-directional flow, but optical flow
should be normalised using the mean vector to help compensate for camera movement. As the motion stream cannot be initialised from image recognition pre-trained models, the authors regularise this stream with multi-task learning. 

Besides, researchers also tend to approximate optical flow using neural networks. Fan et al. \cite{fan2018end} learn motion representations by unrolling a fixed number of optical flow algorithm iterations as layers within a CNN. The authors derive this layer directly from the optimisation of the TV-L1 algortihm with gradient computations being substituted for convolutions with specific kernel instantiations and bicubic warping substituted for bilinear interpolation. The authors go further by relaxing the values of the convolutional kernels to be parameters that are discriminitvely trained using the combined loss of reconstructing TV-L1 optical flow as well as the action recognition loss of the downstream task. Piergiovanni and Ryoo \cite{piergiovanni2019representation} extend this method in two ways. Firstly, the algorithm hyperparameters, such as for the regularisation term and smoothness control, are also learnt discriminatively and shared across layers. Secondly, the optical flow blocks are hierarchically stacked to compute the flow-of-flow which is an abstract representation of the optical flow of similar motion patterns across frames. This is not possible for traditional optical flow as flow representations violate the brightness consistency constraint.
In addition, some methods learn optical flow within a single-stream architecture by utilising a two-stream architecture as supervision during training. Gao et al. \cite{gao2018im2flow} use unsupervised learning to train a CNN to estimate optical flow on previously unseen images. The output of this CNN is used by the motion stream of a two-stream architecture. It then avoids the computation of optical flow during testing, resulting in significantly faster inference. 
Crasto et al. \cite{crasto2019mars} distil pre-trained motion stream features with optical flow input to an untrained appearance stream network with RGB input. The last layer of the network is then fine-tuned with the supervised task loss resulting in an emulation of the motion stream. The authors also train with both losses yielding a motion augmented appearance stream subsuming the two-stream architecture into a single CNN. Stroud et al. \cite{stroud2020d3d} extend this method to 3D CNNs in order to investigate whether 3D convolutions are able to completely subsume motion representations as their spatio-temporal nature would suggest. S3D-G \cite{xie2018rethinking} is used as a baseline architecture trained with the same distillation to which is added an optical flow decoder to measure the presence of optical flow information within the latent features of early network layers. 
In particular, the authors found that
improving motion representations does not improve performance, but improving action recognition performance improves latent motion representations.

\subsection{Motion Feature-based Methods}
Motion feature-based methods mainly learn motion features based on implicit motion cues present in compressed video. For example, Wu et al. \cite{wu2018compressed} find that action recognition methods can be confounded by the excess of information contained within videos and propose to apply CNNs on compressed videos. Three types of coding present in compressed videos are modelled together. 
The authors accumulate residuals and motion vectors such that each of these predictive frames depends only on the intra-frame coding and separate CNN streams can be learnt for each type of coding. This approach lowers computational costs and has a reduced chance of overfitting.
This method also enables learning motion representations without computing optical flow (although it is shown to be orthogonal).
However, fine grain motion details are lost due to the accumulated residuals being blocked together and the accompanying motion vectors being very low resolution. To address this issue, Shou et al. \cite{shou2019dmc} learn motion representations using both motion vectors and accumulated residuals, together, as they are strongly correlated and contain complementary information. This is performed by a lightweight CNN architecture of densely connected layers \cite{huang2017densely} with a skip connection for the motion vector input such that the CNN learns a refinement to motion vectors which are already an approximation to optical flow. The network is trained using additional reconstruction and adversarial losses with both being computed using optical flow which is then not needed for inference.

Another significant body of work dispenses with optical flow and encode motion representations within the appearance stream of single-stream architectures. Sun et al. \cite{sun2018optical} represent motion by applying optical flow inspired operations to the feature maps of 2D CNNs separated over time. This is achieved by an optical flow guided feature network which is trained separately and comprises OFF blocks interleaved with residual blocks. 
Lee et al. \cite{lee2018motion} propose a similar block which pairs temporally adjacent feature maps, shifts one of them in each of the 8 cardinal pixel directions, subtracts these shifts from the other, and aggregates each of these together with the original features. 
These blocks are directly interleaved with the appearance blocks. 
Such one pixel shift is capable of capturing expressive motion due to the much larger and varying size of the receptive field in latent space relative to the input space. 
Mac et al. \cite{mac2019learning} propose another similar block but which has the advantage of being adaptive to the given input by utilising deformable convolutions \cite{dai2017deformable}. Deformable convolutions can learn adaptive positional offsets for their receptive fields. The authors use this deformable receptive field as approximate key point detection. They hypothesise that if these deformation offsets pertain to the same locality across frames, then motion representations can be computed using the change in the vectors implied by the deformation offsets of temporally adjacent frames. 
Moreover, Jiang et al. \cite{jiang2019stm} propose a motion encoding module which uses point convolution to distribute the channel dimension over a number of branches equal to the number of frames in the temporal dimensions. Each branch performs a spatial convolution which is point-wise subtracted from the original features of the temporally previous branch with all branches being concatenated and point convolved back to the starting coordinate space. 

\subsection{Discussion}
In computer vision, optical flow is a well-developed feature to represent the movements of visual contents in local areas and neighboring frames. Incorporating optical flow in action recognition is straight-forward and promising, but the extraction of optical flow requires careful designs. Traditional ways may not carry high-level abstract information to describe motions accurately, while deep learning-based optical flow computation requires mindful pre-training or approximation. Sometimes the computation is also costly. To this end, many recent studies tend to introduce specifically designed deep learning architectures to extract motion information within feature representations implicitly. Further, these motion feature-based methods can be directly optimised with the loss function of base action recognition methods. 
By comparing both types of methods, however, optical flow still attracts significant interests from researchers, thanks to its well-established theories and applications.

\section{Context Representation-based Methods}
\label{sec:ctx}
Context representation usually refers to additional information that is beneficial for the recognition on local instances. 
It has been widely agreed in computer vision that context representation is usually helpful for improving performance.
According to our survey, context representation methods exist in two subcategories (shown in table \ref{tab:subcatcontext}) based on the level of abstraction that the incorporated context is sourced from. The first subcategory of methods uses feature volume contexts which represent the contexts of a single feature derived from the rest of the features in the volume and can be considered global at the sample level. The second subcategory pertains to video context methods which represent the context for the current sample using global video features outside of the sample. Table \ref{tab:subcatcontext} shows some typical methods of this category.

\begin{table}
    \begin{center}
    \resizebox{\columnwidth}{!}{%
    \begin{tabular}{rl}
        \hline
        \textbf{Feature Volume Contexts} & \textbf{Main Contributions}\\
        \hline
        NL \cite{wang2018non} & Global self-attention\\
        GCN \cite{wang2018videos} & Graph convolution over grouped features\\
        TimeCeption \cite{hussein2019timeception} & Large temporal kernel\\
        V4D \cite{zhang2020v4d} & Convolution over clip dimension\\
        \hline
        \textbf{Video Contexts} & \textbf{Main Contributions}\\
        \hline
        AdaScan \cite{kar2017adascan,wu2019liteeval} & MLP frame selector\\
        AdaFrame \cite{wu2019adaframe} & RL LSTM frame selector\\
        \hline
    \end{tabular}
    }
    \end{center}
    \caption{Subcategories and summary of representative context representation methods.}
    \label{tab:subcatcontext}
\end{table}

\subsection{Feature Volume Contexts}
Methods which use feature volume contexts can be further described by three main groups.
Attention-based methods attempt to represent abstract sub-actions earlier in the network by exceeding the local constraints of the CNN kernel with attention mechanisms.
In addition, graph-based methods enable the composition of sub-actions into complex actions by explicitly modelling groups of features using group representation. Other methods enable the the modelling of the global feature volume context with novel classification heads such as by simply using multi-scale temporal resolutions with global receptivity.

Attention-based methods utilise discriminatively learnt attention to facilitate global connectivity within the feature volume. Wang et al. \cite{wang2018non}, inspired by the success of recent self attention models for natural language processing and the denoising method of non-local means for computer vision, proposed the non-local block for modelling long-term spatio-temporal feature volume context. The proposed block is a generic abstraction of self attention and, for each feature, aggregates its signal with the average of all others in the feature volume weighted by the similarity or relationship to its signal. 
Chen et al. \cite{chen20182} propose a double attention module which is a special case of the non-local block that preserves more information by using bilinear pooling, instead of an embedded Gaussian, as its relationship function. Bilinear pooling computes the outer product of the two module inputs which are a dimensionality reduced feature map and a softmax attention map computed with the same features. Another branch of separate attention masks then adaptively routes these global features to the feature in the input they pertain to. Liu et al. \cite{liu2019learning} enrich non-local blocks with a semantic space learnt by an MLP which represents the displacements of the top $k$ inter-frame spatio-temporal correspondences. Diba et al. \cite{diba2018spatio} propose a residual block to model global channel correlations by spatial and spatio-temporal pooling for spatial and temporal branches, respectively. Both branch outputs are processed by fully connected layers terminating in sigmoid functions yielding channel descriptors used for rescaling each input channel rather than full attention maps. Instead of learning adaptive attention maps, Qiu et al. \cite{qiu2019learning} represent the feature volume context in a stream parallel to CNN features with interactions between these global and local representations performed at each block and final representations being classified together by a novel kernel-based classifier. These interactions are residual connections with the signal first undergoing linear projections after global average pooling or upsampling for local to global and global to local connections, respectively.

Another group of methods utilise global connections between groups of features rather than between individual features. Wang and Gupta \cite{wang2018videos} proposed an action recognition method that utilises context representation at the object level rather than the feature level. Object level contexts are extracted based on bounding boxes which are predicted to contain an object in each box. Features of each detected box 
undergo discriminatively learnt transformations to represent their relationships with other objects. The transformed features are then used to compute three adjacency matrices: pairwise object similarity, forward spatio-temporal proximity, and backward spatio-temporal proximity.
Graph convolutions are applied separately over each of these adjacency matrices to facilitate recognition.
Chen et al. \cite{chen2019graph} propose a module which maps to an interaction space defined by an adjacency matrix. The module is learnt based on dimensionreduction and multiplicative projection matrices. Group convolutions are applied to model relationships between features. 
The same projection matrices, along with dimension expansion, are then used to revert to the original spatio-temporal coordinate space such that these global reasoning features can then be integrated with those from convolutional layers. Zhang et al. \cite{zhang2020v4d} utilise the clip dimension for grouping features into clips or 'action units' and propose 4D convolutions for modelling their relationships. This is implemented as an intermediate layer between feature volumes computed by 3D convolutional networks with shared weights operating independently on action units. These volumes are thus able to represent subactions recognised within each clip and can be reasoned with at a high level by the 4D convolution. To perform inference, the authors extend into a 5th dimension to enable averaging of clip level predictions as the typical clips are subsumed by their 4D convolution operator. 

The last group of methods forego early feature volume context representation and only model contexts at the highest level of CNN abstraction within the classification head. Piergiovanni et al. \cite{piergiovanni2017learning} explicitly model subactions as attention filters. The filters are composed of multiple Gaussians with parameterised frame centre, width and stride (between Gaussians). These attention filters are applied to CNN features to weight their aggregation over time just prior to classification. 
Zhou et al. \cite{zhou2018temporal} compute CNN features for different permutations of frames in ordered groupings of various sizes which are less than the total number of sampled frames. For each respective permutation size, an MLP computes a representation of relationships between frames at that grouping size scale. A further MLP aggregates instances of these frame relationship features which are finally summed to the recognised action. Hussein et al. \cite{hussein2019timeception} also apply multiple varied receptive fields to high level CNN features but implement it with a stack of temporal convolution layers instead. Each of these layers comprises of depth-seperable temporal convolutions of multiple varied kernel sizes. Martinez et al. \cite{martinez2019action} augment action recognition CNN architectures with an auxiliary classification branch that utilises cross-channel max pooling to produce spatio-temporally localised features. This takes advantage of the full feature volume context to separate similar classes using using fine-grain local cues of the explicitly modelled variants per class. 

\subsection{Video Contexts}

Here, we refer video contexts as the information from broader video information beyond that of a given sample of frames. Methods that utilise video contexts can be described by 4 types. 
The first type of video context representation-based methods are those that simply utilise the context of useful inductive biases for understanding videos.
A subsequent group of methods weights the aggregation of sample representations based on the context of the latent representations of other samples. Another group uses various forms of neural network to learn effective sampling patterns based on the entire video contexts. Lastly, a group of methods fully exploits dense sampling to account for the the entire video context to improve performance.

In particular, for the first group of methods, researchers utilise the video meta-contexts to perform sampling based on inductive biases relating to the video format and the task of action recognition. The TSN framework \cite{wang2016temporal} provides an example of where these inductive biases, pertaining to all videos, can be utilised to avoid the computational expense of dense sampling and to achieve efficient coverage of poorly localised actions. The inductive bias in this case is that dense sampling is not commensurate with the semantics of actions where proximal frames contain, on average, vastly more redundant information than distant ones. 
The sparse sampling and video level supervision of the TSN framework is extended by Liu et al. \cite{liu2018t} for 3D CNNs which are able to incorporate video level context within the scope of the temporal receptive field. Temporal interactions across sparsely sampled frames combined with video level supervision enables learning features with some semblance of respect to the consistent sampling of frames that are unlikely to be proximal. Zolfaghari et al. \cite{zolfaghari2018eco} propose an architecture that improves on the efficiency of this method by stacking a 2D CNN with subsequent parallel processing by separate 2D and 3D CNN streams which enables the 3D CNN to reserve its capacity for learning high level video context features.

Subsequently, there are a group of methods which eschews computational efficiencies in favour of resolving frame importance in latent space. The latent space is not used retroactively for sampling but is used for weighted pooling of latent frame representations. Wang et al. \cite{wang2017untrimmednets} augment a two-stream framework with two key modules to enable informed pooling of clip level features by utilising global video information. The first module is a clip sampler which, rather than uniformly sampling clips, samples clips with respect to the start and end of different shots. 
The second module learns to weight clip level representations discrimitively using either soft attention mechanisms to weight all clip level class scores or using multiple instance learning such that only the features of the most discrimitive clips are aggregated enabling more coverage for poorly localised actions. Kar et al. \cite{kar2017adascan} propose another adaptive pooling method. Given a set of frame level features extracted by a CNN, this method uses an MLP that is iteratively fed each frame level representation as well as the current pooled representation. 
The MLP outputs an importance score for the current frame is used to rescale both the pooled representation and the current frame which are then aggregated. A regularising loss is further introduced to encourage sparse frame selection based on the inductive bias that most actions are identifiable by very few frames.

In another group of studies, researchers 
utilise the video contexts to help identify which frames ought to be sampled for improving both localisation of sparse actions and the potential for computational efficiency. Korbar et al. \cite{korbar2019scsampler} propose to sample clips at test time using saliency scores. The authors use a light-weight CNN to quickly compute action class scores for each clip within the video with the saliency score being the magnitude of the max action class for each clip. Wu et al. \cite{wu2019liteeval} incorporate the sampler implicitly by learning whether to compute a fine-grain network's computationally expensive features. The decision is made depending on previous hidden state and the current input whilst computing coarse-grain features for all frames with light weight network. Both the fine grain network and the coarse grain network are instantiated by CNN-LSTMs. The decision to compute the expensive features is learnt by an MLP with optimisation being differentiable by a reparameterisation of the categorical distribution given by the binary output of the MLP. In another study, Wu et al. \cite{wu2019adaframe} utilise a reinforcement learning to perform frame sampling. They feed the features of the backbone CNN to a reinforcement learning `agent' LSTM with three output branches, including a frame selection policy, a classification of the video, and a prediction of how helpful sampling the next frame will be. The `agent' LSTM is supplemented by a very light weight CNN applied to the entire downsampled video whose features are attention queried with the previous time-step output of the `agent' LSTM, so that the most relevant parts of the global video context can be attended.

The last group of methods we review take a very different approach in which researchers perform computationally burdensome dense sampling in order to provide the full spatio-temporal feature volume contexts when recognising actions in sampled portions of video. Bhardwaj et al. \cite{bhardwaj2019efficient} attempt to preserve efficiency by only performing dense sampling during training where they disti the global video contexts learnt by a teacher network, trained on every video frame, to a student network which only sees a factor of those frames. 
Wu et al. \cite{wu2019long} aggregate clip features extracted using dense sampling over a wide temporal window, yielding a feature bank which is then queried using short term features centred on this window for multi-class action recognition on untrimmed videos. This enables relating the objects and subjects in the current scene to the longer term context. 

\subsection{Discussion}

Each subcategory of context representation-based methods elicits advantages but at commensurate computational costs. Feature volume context-based methods hierarchically augment features with global contexts and learn these augmentative connections discriminitively with the target task of action recognition. Computationally, these methods add little to the overall cost of underlying backbone architectures. The feature volume contexts deliver rich information for accurate recognition, especially combining with broad random sampling of many clips during inference. Video context methods require additional computing to extract features from the entire video, but this is sometimes offset by using lightweight networks or fewer-frame approximations to the video context. They are able to access information that other methods may not have access to.
Some of these methods directly optimise frame sampling which is a fundamental procedure in action recognition and enables much finer control over problems like representation bias where biased cues can be avoided for more robust representation. 

\section{Experimental Results}
\label{sec:exp}
In this study, we conduct comprehensive quantitative analysis on the popular algorithms we have reviewed. We will list the detailed results of compared methods in this section.


\subsection{Evaluation Methods}
There are many action recognition datasets that provide effective evaluation on various methods. Action recognition datasets pose various challenges based on the properties of their constituent videos and actions classes. Action recognition methods are typically evaluated by computing the quantitative performance on a predefined test or validation set using a predefined training set. Mutually agreed upon benchmark datasets facilitate comparison between methods and can be targeted towards measuring whether each method addresses specific challenges. 
The nature of datasets often determines the metrics by which quantitative performance is measured.


Action recognition datasets pose various challenges to the methods evaluated in this section, but dataset selection mediates the challenge of representing motion insofar as methods that address the challenge improve performance much more on some datasets than others. This can also be the case for datasets where temporal dynamics are emphasised and where temporal or spatio-temporal representations significantly improve performance. The sparse action challenge is pertinent for all datasets with some featuring specifically untrimmed videos but, even for those with predominantly trimmed videos, the granularity of the trimming may not be sufficient to perfectly localise the action. With a small granularity, samples will consistently retain blocks of frames that will not represent the action class. Many datasets feature complex action classes with some being solely comprised of them. Some benchmark datasets completely overlook the challenges of real world action recognition relying solely on powerful scene-based representations and other forms of representation bias which voids many possible applications and emphasises spatial representations. 

In order to comprehensively illustrate the effectiveness of action recognition algorithms, we introduce their performance on various datasets. The datasets that feature for the methods reviewed in this article and which we use to compare them include:
\begin{itemize}
    \item[] \textbf{UCF101}\cite{soomro2012ucf101}: 101 action classes. 9537 and 3783 training and testing samples, respectively, with an average duration of 7.21 seconds each. Representation biased emphasised by small dataset size.
    \item[] \textbf{HMDB51}\cite{kuehne2011hmdb}:	51 action classes. 6766 total samples. Representation biased due to small dataset size. 
    \item[] \textbf{Sports1M}\cite{karpathy2014large}: Over 1 million total samples. 487 sports classes with strong representation bias owing to the consistency of scene correlations in sports-based videos. 
    \item[] \textbf{Kinetics400/600}\cite{kay2017kinetics,carreira2017quo,carreira2018short}: 306245/495547 total samples of approximately 10 second duration. 400/600 action classes with representation bias offset by large dataset size and diversity. 
    \item[] \textbf{Something-Something V1/2 (SS-V1/2)}\cite{goyal2017something}: 108499/220847 total samples of approximately 4 second average duration. 174 classes for both versions with an emphasis on temporal dynamics with, for example, the arrow of time being an essential feature for some action classes. This temporal emphasis is achieved by all actions being object agnostic and strong inter-class similarities e.g. `moving something and something so they pass each other' and `moving something closer to something'.
    \item[] \textbf{Charades}\cite{sigurdsson2018charades}: 7985 training samples of 30.1 second average duration. 157 action classes featuring realistic depictions of complex everyday actions. 
    \item[] \textbf{Epic Kitchens}\cite{damen2018scaling}: Egotistical dataset of 39596 total samples focusing on fine-grain variations of both object and action. As such, this dataset comprises metrics for recognition of the object (noun), object agnostic action (verb) and the object-aware action resulting from the possible combinations of noun and verb. There are two main metrics for evaluation seen and unseen. Seen is where the same kitchens feature for training and testing and unseen is where only new kitchens are featured in the test set.
    \item[] \textbf{Moments in Time (MiT)}\cite{monfort2019moments}: Over 1 million total samples of approximately 3 second duration each. 339 action classes noting that this dataset is not strictly for action recognition as per our definition as some of the action classes are not performed by humans but, with these in the vast minority, we include results for this dataset for completeness.
    \item[] \textbf{ActivityNet 1.2/1.3}\cite{caba2015activitynet}: 100/200 classes and approximately 13000/28000 total samples. With 1365.5 average frames per sample, this dataset specifically features untrimmed videos. 
    \item[] \textbf{Diving48}\cite{li2018resound}: 16067 training and 2,337 test samples with an average of 159.6 frames each. Very little representation bias due to all samples of all 48 classes having the almost identical background of a competition diving pool.
\end{itemize}

To facilitate a fair comparison of surveyed methods, performance is given using widely used metrics and mediating factors are also taken into consideration for our analysis. 
Performance is evaluated in either accuracy or mean average precision (mAP), the latter of which is the average of precision at each recall value averaged over each class. \textbf{Charades} and both versions of \textbf{ActivityNet} are multi-class datasets with performance being reported in mAP. All other datasets, herein, are single action and reported only in accuracy. 
Although performance is often reported for clip and video level recognition, we do not consider clip level performance as a meaningful metric, especially, considering that it would lead to bias in favour of methods that process entire videos as a clip. The performance metric of top 5 accuracy is not included here for both clarity and brevity. 
Another factor to be considered for a fair representation of relative performance is the extent that auxiliary data is used for pre-training, or otherwise, and modalities that are not sourced from RGB frames such as audio. It is the most descriptive of how methods vary within our proposed categorisation framework. 
Lastly, although action recognition performance is also mediated by computational costs, there is not a unified hardware and software platform to implement all the compared methods. We omit the comparison of computational speeds in this paper. 

\subsection{Results}
\begin{table*}
    \begin{center}
    \resizebox{\linewidth}{!}{%
    \begin{tabular}{rlllllllll}
        & \textbf{Modality} & \textbf{UCF101} & \textbf{HMDB51} & \textbf{Sports1M} & \textbf{Kin400} & \textbf{Kin600} & \textbf{SSV1} & \textbf{SSV2}  & \textbf{Charades}\\
        \hline
        \multicolumn{10}{l}{\textbf{2D Convolutional Methods}}\\
        \hline
        Fusion CNNs \cite{karpathy2014large} & RGB & 65.4 & & 63.9 & & & & & \\ 
        Donahue et al. \cite{donahue2015long} & RGB & 82.66 & & & & & & & \\
        Yue-Hei Ng et al. \cite{yue2015beyond} & RGB+Flow & 88.6 & 73.1 & & & & & & \\
        Dynmic Image Net \cite{bilen2016dynamic} & RGB+Flow+IDT & 96.0 & 74.9 & & & & & & \\
        TSN \cite{wang2016temporal} & RGB+Flow+Warped & 94.2 & 69.4 & & & & & & \\
        Cool-TSN \cite{roberto2017procedural} & RGB+Flow & 94.2 & 69.5 & & & & & \\
        ST-ResNet (Multiplier) \cite{feichtenhofer2017spatiotemporal} & RGB+Flow+IDT & 94.9 & 72.2 & & & & & & \\
        STCB \cite{wang2017spatiotemporal} & RGB+Flow & 94.6 & 68.9 & & & & & & \\
        ActionVLAD \cite{girdhar2017actionvlad} & RGB+IDT & 93.6 & 69.8 & & & & & & 21.0 mAP\\ 
        $L^2$STM \cite{sun2017lattice} & RGB & 93.6 & 66.2 & & & & & & \\
        SVMP \cite{wang2018video} & RGB+IDT & & 81.3 & & & & & & 26.7 mAP\\
        CO2FI \cite{lin2018action} & RGB+Flow+IDT & 95.2 & 72.6 & & & & & & \\
        TSM \cite{lin2019tsm} & RGB+Flow & & & & & & 52.6 & 66.0 & \\
        bLVNet \cite{fan2019more} & RGB+Flow & & & & & & & 68.5 & \\
        GSM \cite{sudhakaran2020gate} & RGB & & & & & & 55.16 & & \\
        \hline
        \multicolumn{10}{l}{\textbf{3D Convolutional Methods}}\\
        \hline
        C3D \cite{tran2015learning} & RGB+IDT & 90.4 & & 61.1 & & & & \\
        ST-ResNet \cite{feichtenhofer2016spatiotemporal} & RGB+Flow+IDT & 94.6 & 70.3 & & & & & & \\ 
        P3D \cite{qiu2017learning} & RGB & 88.6 & & 66.4 & & & & & \\
        MiCT \cite{zhou2018mict} & RGB+Flow & 94.7 & 70.5 & & & & & & \\
        TrajectoryNet \cite{zhao2018trajectory} & RGB & & & & & & 47.8 & & \\
        ARTNet \cite{wang2018appearance} & RGB & 94.3 & 70.9 & & 70.7 & & & & \\ 
        MF-Net \cite{chen2018multi} & RGB & 96.0 & 74.6 & & 72.8 & & & & \\
        S3D-G \cite{xie2018rethinking} & RGB & 96.8 & 75.9 & & 74.7 & & 48.2 & & \\
        R(2+1)D \cite{tran2018closer} & RGB+Flow & 97.3 & 78.7 & 73.0 & & & & & \\
        StNet \cite{he2019stnet} & RGB & 95.7 & & & 71.38 & 78.99 & & & \\
        AssembleNet \cite{ryoo2019assemblenet} & RGB+Flow & & & & & & & & 58.6 mAP \\
        GST \cite{luo2019grouped} & RGB & & & & & & 48.6 & 62.6 & \\
        SlowFast \cite{feichtenhofer2019slowfast} & RGB & & & & 79.8 & 81.8 & & & 45.2 mAP\\
        CoST \cite{li2019collaborative} & RGB & & & & 77.5 & & & & \\
        CSN \cite{tran2019video} & RGB & & & 75.5 & 82.6 & & 53.3 & & \\
        RMS \cite{kim2020regularization} & RGB & & & & 76.3 & & & 61.2 & \\
        I3D \cite{carreira2017quo} & RGB+Flow & 98.0 & 80.9 & & 74.2 & & & & \\
        Wei et al. \cite{wei2018learning} & RGB & 88.2 & 55.4 & & & & & & \\
        IG-Kinetics \cite{ghadiyaram2019large} & RGB & & & & 82.8 & & 51.6 & & \\
        Korbar et al. \cite{korbar2018cooperative} & RGB+Audio & 83.7 & 53.0 & & & & & & \\
        Xu et al. \cite{xu2019self} & RGB & 72.4 & 30.9 & & & & & & \\
        Kim et al. \cite{kim2019self} & RGB & 65.8 & 33.7 & & & & & & \\ 
        Wang et al. \cite{wang2019self} & RGB & 61.2 & 33.4 & & & & & & \\
        \hline
        \multicolumn{10}{l}{\textbf{Motion Representation Methods}}\\
        \hline
        Simonyan and Zisserman \cite{simonyan2014two} & RGB+Flow & 88.0 & 59.4 & & & & & & \\
        MFNet \cite{lee2018motion} & RGB & & & & & & 43.92 & & \\ 
        OFF \cite{sun2018optical} & RGB+RGB Diff & 96.0 & 74.2 & & & & & & \\
        Fan et al. \cite{fan2018end} & RGB+IDT & 95.4 & 72.6 & & & & & & \\
        CoViAR \cite{wu2018compressed} & RGB & 90.4 & 59.1 & & & & & & 21.9 mAP\\
        STM \cite{jiang2019stm} & RGB & 96.2 & 72.2 & & 73.7 & & 50.7 & 64.2 & \\
        FcF \cite{piergiovanni2019representation} & RGB & & 81.1 & & 77.9 & & & & \\
        DMC-Net \cite{shou2019dmc} & RGB & 96.5 & 77.8 & & & & & & \\
        MARS \cite{crasto2019mars} & RGB & 97.6 & 79.5 & & 74.8 & & 51.7 & & \\
        D3D \cite{stroud2020d3d} & RGB & 97.6 & 80.5 & & 76.5 & 79.1 & & & \\
        \hline
        \multicolumn{10}{l}{\textbf{Context Representation Methods}}\\
        \hline
        Piergiovanni et al. \cite{piergiovanni2017learning} & RGB & & 68.0 & & & & & & \\
        NL \cite{wang2018non} & RGB & & & & 77.7 & & & & 39.5 mAP\\ 
        STC \cite{diba2018spatio} & RGB & 95.8 & 72.6 & & 68.7 & & & & \\
        GCN \cite{wang2018videos} & RGB & & & & & & 46.1 & & 39.7 mAP\\
        $A^2$-Net \cite{chen20182} & RGB & 96.4 & & & 74.6 & & & & \\
        TRN \cite{zhou2018temporal} & RGB+Flow & & & & & & 42.01 & 55.52 & 25.2 mAP\\ 
        GloRe \cite{chen2019graph} & RGB & & & & 76.09 & & & & \\
        Timeception \cite{hussein2019timeception} & RGB & & & & & & & & 41.1 mAP\\ 
        Martinez et al. \cite{martinez2019action} & RGB & & & & 78.8 & & 53.4 & & \\
        CPNet \cite{liu2019learning} & RGB & & & & 75.3 & & & 57.65 & \\ 
        LGD \cite{qiu2019learning} & RGB+Flow & 98.2 & 80.5 & & 81.2 & 83.1 & & & \\
        V4D \cite{zhang2020v4d} & RGB & & & & 77.4 & & 50.4 & \\
        T-C3D \cite{liu2018t} & RGB & 92.5 & 62.4 & & & & & & \\
        ECO \cite{zolfaghari2018eco} & RGB & 94.8 & 72.4 & & 70.0 & & 46.4 & & \\
        SCSampler \cite{korbar2019scsampler} & RGB & & & 84.0 & 80.2 & & & & \\
        LFB \cite{wu2019long} & RGB & & & & & & & & 42.5 mAP\\ 
        \hline
    \end{tabular}
    }
    \end{center}
    \caption{Quantitative results, as a percentage and in accuracy, unless otherwise noted.}
    \label{tab:main}
\end{table*}

\begin{table}
    \begin{center}
    \begin{tabular}{rlll}
        \hline
        & \textbf{Version} & \textbf{Modality} & \textbf{mAP}\\
        \hline
        UntrimmedNet \cite{wang2017untrimmednets} & 1.2 & RGB+Flow & 91.3\\
        AdaFrame \cite{wu2019adaframe} & 1.3 & RGB & 71.5\\
        LiteEval \cite{wu2019liteeval} & 1.3 & RGB & 72.7\\
        P3D \cite{qiu2017learning} & 1.3 & RGB & 78.9\\
        V4D \cite{zhang2020v4d} & 1.3 & RGB & 88.9\\
        \hline
    \end{tabular}
    \end{center}
    \caption{Quantitative results for the \textbf{ActivityNet} dataset. Results are given as a percentage and are for test and validation sets for version 1.2 and 1.3, respectively.}
    \label{tab:an}
\end{table}

\begin{table}
    \begin{center}
    \begin{tabular}{rll}
        \hline
        & \textbf{Modality} & \textbf{Accuracy}\\
        \hline
        GST \cite{luo2019grouped} & RGB & 38.8\\
        GSM \cite{sudhakaran2020gate} & RGB & 40.27\\
        \hline
    \end{tabular}
    \end{center}
    \caption{Results for the \textbf{Diving48} dataset given as a percentage and for the test set.}
    \label{tab:d48}
\end{table}

\begin{table}
    \begin{center}
    \begin{tabular}{rll}
        \hline
        & \textbf{Modality} & \textbf{Accuracy}\\
        \hline
        bLVNet \cite{fan2019more} & RGB & 31.4\\ 
        AssembleNet \cite{ryoo2019assemblenet} & RGB+Flow & 34.27\\
        CoST \cite{li2019collaborative} & RGB & 32.4\\
        \hline
    \end{tabular}
    \end{center}
    \caption{Results for the \textbf{Moments in Time} dataset given as a percentage and for the validation set.}
    \label{tab:mit}
\end{table}

\begin{table}
    \begin{center}
    \resizebox{\columnwidth}{!}{%
    \begin{tabular}{rllllllllll}
        \hline
        & \textbf{Modality} & \multicolumn{3}{c}{\textbf{Seen}} & \multicolumn{3}{c}{\textbf{Unseen}} & \multicolumn{3}{c}{\textbf{Val}}\\
        & & V & N & A & V & N & A & V & N & A \\
        \hline
        LFB \cite{wu2019long} & RGB & 60.0 & 45.0 & 32.7 & 50.9 & 31.5 & 21.2 & 53.0 & 31.8 & 22.8\\
        LSTA \cite{sudhakaran2019lsta} & RGB & & & 30.16 & & &15.88 & & & \\
        GSM \cite{sudhakaran2020gate} & RGB & 59.41 & 41.83 & 33.45 & 48.28 & 26.15 & 20.18 & & & \\
        GST \cite{luo2019grouped} & RGB & & & & & & & 56.50 & & \\
        IG-Kinetics \cite{ghadiyaram2019large} & RGB & 65.2 & 45.1 & 34.5 & 57.3 & 35.7 & 25.6 & 58.4 & 36.9 & 26.1 \\
        \hline
    \end{tabular}
    }
    \end{center}
    \caption{Quantitative results for the \textbf{Epic Kitchens} dataset featuring both modes of test set evaluation as well as validation set results. Performance is given, as a percentage and in accuracy, for each task.}
    \label{tab:ek}
\end{table}

The results for the main contingent of competitive benchmarks is featured in table \ref{tab:main}. 
Note that we mainly compare methods according to their validation results. This is because the validation results can cover related methods more comprehensively. Besides, we do not have access to test labels of included datasets. For
\textbf{UCF101} and \textbf{HMDB51}, the related results are in average accuracy over three predefined splits. 
Table \ref{tab:an} features results for the untrimmed video dataset \textbf{ActivityNet}. Noting that this table features context representation methods in the majority. Tables \ref{tab:d48}, \ref{tab:mit} and \ref{tab:ek} feature more difficult and recent action recognition benchmarks that have less current results. Note that some datasets use online videos which are occasionally removed resulting in methods being evaluated on different datasets depending on their publication dates.

According to the results, we can roughly observe that 3D convolutional methods usually achieve better performance than 2D convolutional methods. 
We can also find that self-supervised learning methods achieve poorer performance, but this is merely a reflection of the intended purpose which is to attempt to substitute for pre-training thereby reducing the need for additional data. 
Note that most motion representation methods do not use optical flow input modality in table \ref{tab:main} and that the modality computed by these methods is often directly compatible with the other methods featured in this table. The apparent shortcomings of motion representation methods is further exacerbated by their primary motivation being efficiency with respect to two-stream architectures. Although not featured in table \ref{tab:main}, many motion representation methods obtain performance very close or in excess of those obtained by using an equivalent architecture or optical flow instead of their learnt approximation. 
We can also observe that in recent years there has been a stark reduction in the usage of the optical flow modality which may be due to its very prohibitive impact on inference time. Although recent motion representation methods are able to replicate or exceed the performance gained from the optical flow modality, their uptake remains low across all benchmarks which may simply be a reflection of researchers seeking to avoid conflating the source of performance gains in their experiments.

Another important accompaniment to the results in table \ref{tab:main} is that some methods outside of the feature volume context category utilise backbone networks which include those methods \cite{wang2018non}. This highlights the ubiquity of feature volume context representation methods and the orthogonality of the performance improvements they garner with most other methods being improved by their inclusion. Considering the very minor increase in computational costs, it seems clear that some forms of feature volume contexts should be used for action recognition. Although most video context representation methods are applied directly to untrimmed videos of the conventional sense, the performance of other video methods in table \ref{tab:main} show that the video context is useful, even for trimmed clips.

The accuracy of modern methods on smaller scale datasets such as \textbf{UCF101} and \textbf{HMDB51} has reached a bottleneck such that they don't serve as good performance indicators. Given their small scale, it is uncertain whether the slightly better performing methods don't simply use representation biases which can be very powerful on such a small scale but do not transfer to downstream tasks. \textbf{Diving48} presents an antithesis in terms of representation bias exhibiting almost none due to the common features exhibited in all samples for all of its action classes. Table \ref{tab:d48} shows that only very recent methods are applied to this dataset and that even their results are poor. The results on \textbf{Epic Kitchens} is perhaps a more explicit example of action recognition methods using representation biases with a large discrepancy between seen and unseen kitchen performance for all methods in table \ref{tab:ek}. Representation bias is of further concern considering the much lower diversity of \textbf{Something-Something} compared with \textbf{Kinetics} and yet the latter experiences much higher validation accuracy for the same methods despite the ratio of samples to classes being roughly consistent between the two. 

In table \ref{tab:main} it is clear that temporal shifting is a very successful method for temporal representation learning especially on the temporally focused \textbf{Something-Something} datasets with strong performance even compared with 3D methods. This is further underscored by its simplicity and low computational cost. Other than temporal shifting methods, 3D convolutional methods vastly outperform 2D methods, especially on the large-scale \textbf{Kinetics} datasets. LSTMs do not appear to work well as the primary means of temporal representation learning but do perform well in auxiliary functions such as frame sampling for video context representation methods \cite{wu2019adaframe}. Otherwise, There is comparable performance achieved by methods from each category within our taxonomy demonstrating the importance of each.


\subsection{Looking Forward}

Here we outline some avenues for future research directions that could be important based on our categorisation and discussion of action recognition methods.

Firstly, there is still much space for further development of the reviewed categories, such as 3D convolutional methods and motion representations. We believe that researchers can continue to push the limit of deep learning architectures for appearance-based action recognition by introducing more accurate models with lower computational costs. In addition, besides the reviewed major categories, we find that the success of shifting channels temporally encourages further investigation.
This could include temporally shifting 3D channels at the frame level like their 2D convolutional counterparts \cite{lin2019tsm}, or at the clip level similar to Zhang et al. \cite{zhang2020v4d}. Existing state-of-the-art shifting methods could also be refined such as by combining the longer range temporal shift of TAM \cite{chen2018biglittle} with the gating of GSM \cite{sudhakaran2020gate} and incorporate shuffling or channel dropout to curtail overfitting in the gating parameters of GSM.

The Kinetics datasets \cite{kay2017kinetics} present an excellent opportunity to test the potential performance of methods that utilise the entire video length as each sample video is a crop from a longer video that is available online. A thorough study about video contexts can be carried out by 
comparing performance against existing methods according to contexts from both the full video and the human derived crops of the initial dataset. This comparison could potentially expose the viability of video context methods and, conversely, the viability of applying existing methods to untrimmed video as would be required by many downstream tasks. The entire video context may be too long to both encompass the essential features throughout the action's temporal extent and distinguish the action from the remaining noise. 

Furthermore, we find that there remains a disconnection between the untrimmed videos of real world applications and trimmed benchmark videos, in terms of supervision. The labelled action may not occur for the full clip duration, depending on the dataset, or may have no semantic consensus (amongst humans) as to what constitutes the start or end of the action. We think that such problem could motivate further structuring of action recognition output and supervisory signal. This could take the form of sequential/hierarchical representations in the output or over supervised action recognition tasks where fine grain labelling, such as those used in action localisation, is used for training on the coarse grain task of action recognition.

   
Since Wu et al. \cite{wu2019long} achieved substantial performance improvements and halved training time by simply tuning the learning rate and iterations of existing training regimes, we believe that current and future advances in image recognition architectures, training regimes and regularisation methods can deliver further improvement on performance of action recognition. A comprehensive study of the transferability of these methods should be undertaken to ascertain trends in this regard.

Lastly, given the many mediating costs of action recognition, including the considerable computational cost incurred by the video medium itself, unification of efficiency related metrics is key to understanding action recognition performance and a standardisation of reporting, perhaps into a single, paramount metric, is an ideal worthy of investigation.

%
%

\ifCLASSOPTIONcaptionsoff
  \newpage
\fi



\bibliographystyle{IEEEtran}
\bibliography{references.bib}
\end{document}